\documentclass[onecolumn,journal,12pt]{IEEEtran}
\IEEEoverridecommandlockouts
\usepackage{mathrsfs}
\usepackage{amssymb}
\usepackage{amsmath}
\usepackage{amssymb}
\usepackage{amsthm}
\usepackage{multirow,makecell}
\usepackage{makecell}
\usepackage[table]{xcolor}
\usepackage{subfigure}
\usepackage{setspace}
\usepackage{stfloats}
\usepackage{graphicx}
\usepackage{epstopdf}
\usepackage{soul}
\soulregister\cite7
\soulregister\ref7
\usepackage{url}
\usepackage{color}
\usepackage{xcolor}

\ifCLASSINFOpdf \else \fi

\usepackage{algorithm}
\usepackage{algorithmic}

\usepackage{cite}

\begin{document}
\title{MmWave Radar and Vision Fusion for Object Detection in~Autonomous Driving: A~Review}
\author{Zhiqing Wei,
		Fengkai Zhang,
		Shuo Chang,
		Yangyang Liu,
		Huici Wu,
		Zhiyong Feng
\thanks{
Zhiqing Wei, Fengkai Zhang, Shuo Chang, Yangyang Liu and Zhiyong Feng
are with Key Laboratory of Universal Wireless Communications, Ministry of
Education, School of Information and Communication Engineering, Beijing
University of Posts and Telecommunications, Beijing, 100876, China (e-mail:
\{weizhiqing, zhangfk, changshuo, yangyangl, fengzy\}@bupt.edu.cn).

Huici Wu is with the National Engineering Lab for Mobile Network
Technologies, Beijing University of Posts and Telecommunications, Beijing
100876, China (e-mail: dailywu@bupt.edu.cn).

Correspondence authors: Zhiqing Wei.}}


\maketitle

\begin{abstract}
With autonomous driving developing in~a~booming stage, accurate object detection in~complex scenarios attract wide attention to ensure the safety of autonomous driving.
Millimeter wave (mmWave) radar and vision fusion is a~mainstream solution for accurate obstacle detection. 
This~article presents a~detailed survey on mmWave radar and vision fusion based obstacle detection methods. 
First, we introduce the tasks, evaluation criteria, and datasets of object detection for autonomous driving.
The process of mmWave radar and vision fusion is then divided into three parts: sensor deployment, sensor calibration, and sensor fusion, which are reviewed comprehensively.
Specifically, we classify the fusion methods into data level, decision level, and feature level fusion methods. 
In addition, we introduce three-dimensional(3D) object detection, the fusion of lidar and vision in~autonomous driving and multimodal information fusion, which are promising for~the~future.
Finally, we summarize this article.

\end{abstract}


\IEEEpeerreviewmaketitle

\section*{GLOSSARY}
\begin{table}[!h]
	\renewcommand{\arraystretch}{1.1}
	\begin{center}
		\begin{tabular}{m{0.1\textwidth}m{0.4\textwidth}}
			2D & Two-dimensional\\
			3D & Three-dimensional\\
			ACC & Autonomous Cruise Control \\
			ADAS & Advanced Driver Assistance System\\
			ADS &Automated Driving System\\
			AI & Artificial Intelligence\\
			ALV &Autonomous Land Vehicle\\
			AP & Average Precision\\
			AR & Average Recall \\
			CMGGAN & Conditional Multi-Generator Generative Adversarial Network \\
			CNNs & Convolutional Neural Networks \\
			DARPA & Defense Advanced Research Projects Agency\\
			DCNN & Deep Convolutional Neural Network \\
			DPM &Deformable Parts Model\\
			EKF & Extended Kalman Filter \\
			FCN & Fully Convolutional Neural Network\\
			FFT & Fast Fourier Transform\\
			FOV	& Field of View \\
			GPS & Global Positioning System \\
			GVF & Gradient Vector Flow\\
			HOG &Histograms of Oriented Gradients\\
			IoT & Internet of Things\\
			IoU & Intersection over Union \\
			
		\end{tabular}
	\end{center}
\end{table}
\begin{table}[!h]
	\renewcommand{\arraystretch}{1.1}
	\begin{center}
		\begin{tabular}{m{0.1\textwidth}m{0.4\textwidth}}

			mAP & Mean Average Precision \\
			MILN & Multilayer In-place Learning Network\\
			mmWave &Millimeter Wave\\
			MTT & Multi-Target Tracking\\
			PR & Precision Recall \\
			ROI & Region of Interest\\
			RPN & Regional Advice Network \\
			SAE &Society of Automotive Engineers\\
			SAF & Spatial Attention Fusion \\
			V2X & Vehicle to Everything\\

		\end{tabular}
	\end{center}
\end{table}

\section{Introduction}

Main causes of traffic accidents are the complicated road conditions and a~variety of human errors. Autonomous driving technology can leave the judgment of road conditions and human manipulation to the vehicle itself, which can improve traffic efficiency and driving safety. In~recent years, with~the significant development of artificial intelligence (AI), Internet of Things (IoT), and~mobile communication, etc., autonomous driving has developed rapidly. There is no doubt that in~the~foreseeable future, autonomous driving vehicles will enter people's daily life and become a~common AI product. The~earliest research on autonomous driving was in~the~1980s when Defense Advanced Research Projects Agency (DARPA) organized the autonomous land vehicle (ALV) project~\cite{001}. In~2009, Google announced the establishment of a~research team to start research on autonomous driving technology~\cite{002}. In~2013, Baidu launched research on “baidu nomancar”. In~the~same year, Audi, Nissan, and other traditional automobile manufacturers began to layout autonomous driving cars. {{However}, until~now, autonomous driving has not fully entered people's life. The~complex traffic environment has too many uncertainties, which discourage people from entrusting their safety to autonomous driving vehicles.}

In 2014, the Society of Automotive Engineers (SAE) released j3016 standard, which defined the standards of autonomous driving level from L0 (no driving automation) to L5 (full~driving automation) \cite{004}. Levels L4 and L5 can completely get rid of manual operation during driving. However, there are more challenges to guarantee the safety of driving~\cite{003}. For~safety reasons, the~anti-collision function is an essential part of automated driving system (ADS), and~the core task of the anti-collision system is obstacle detection. Therefore, one of the challenges for high-level autonomous vehicles is accurate object detection in~complex~scenarios.

Object detection is the most popular research area in~the~field of vision image processing. In~the~early stage of research, due to the lack of efficient image feature representation methods, object detection algorithms relied on the construction of manual features. Representative traditional algorithms included histograms of oriented gradients (HOG) \cite{005}, deformable parts model (DPM) \cite{006}, and so on. In~recent years, as~deep learning has developed, its great potential in~object detection has been discovered. Detection algorithms based on deep learning are divided into two types: one-stage and two-stage. The~former contains YOLO~\cite{007}, SSD~\cite{008}, and RetinaNet~\cite{009}, and~the latter contains R-CNN~\cite{010}, Fast R-CNN~\cite{011}, and Faster R-CNN~\cite{012}, etc. However, the~current visual object detection algorithms have hit a~performance ceiling because the detection algorithms face very complicated situations in~practice. For~the scenarios of autonomous driving, the~obstacles include pedestrians, cars, trucks, bicycles, and~motorcycles, and~obstacles within the visual range have different scales and length-to-width ratios. In~addition, there may be varying degrees of occlusion between obstacles, and~the appearance of the obstacles may be blurred because of the extreme weather such as rainstorms, heavy snow, and fog, which result in~a~great reduction in~detection performance~\cite{013}. Studies have shown that CNN has poor generalization ability for untrained distortion types~\cite{014}. 

{{Various} studies have shown that a camera is not sufficient for autonomous driving tasks.} Compared with vision sensors, the~detection performance of millimeter wave (mmWave) radar is less affected by extreme weather~\cite{015,016}. In~addition, the~mmWave radar not only measures the distance, but~also uses the Doppler effect of the signal reflected by the moving object to measure the velocity vector~\cite{017,018}. However, mmWave radar cannot provide the outline information of the object, and~it is difficult to distinguish relatively stationary targets. In~summary, the~detection capabilities of the vision sensor and the mmWave radar could complement each other. The~detection algorithms based on mmWave radar and vision fusion can significantly improve the perception of autonomous vehicles and help vehicles better cope with the challenge of accurate object detection in~complex~scenarios.

\begin{figure}[H]
	\includegraphics[width=1.0\textwidth]{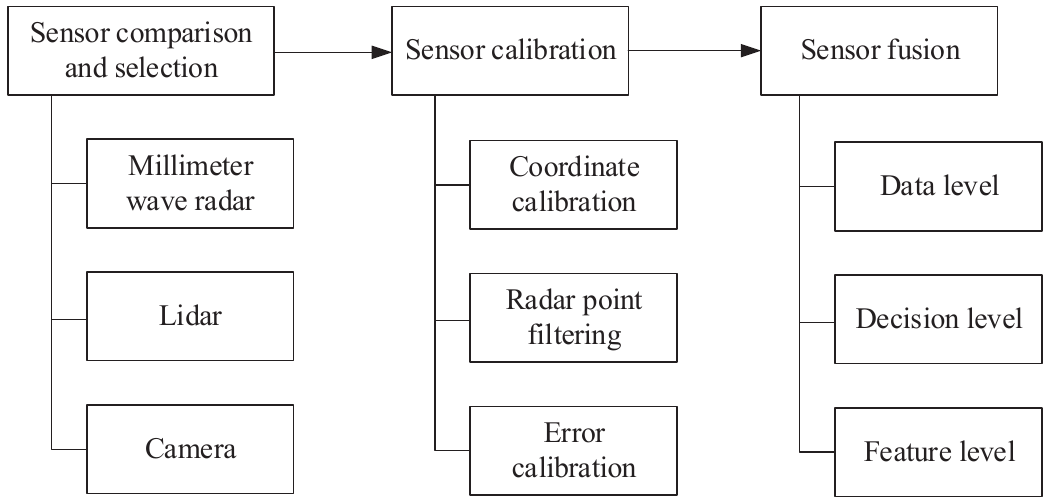}
	\caption{Process of mmWave radar and vision~fusion.}
	\label{Process of mmWave radar and vision fusion}
\end{figure}

The process of mmWave radar and vision fusion based object detection is shown in~Figure~\ref{Process of mmWave radar and vision fusion}. The~processes of mmWave radar and vision fusion consist of three parts: sensor selection, sensor calibration, and sensor fusion. In~order to achieve the desired performance of object detection with mmWave radar and vision fusion, the~following challenges need to be~solved.
\begin{itemize}
	
	\item Spatiotemporal calibration: The~premise of fusion is to be in~the~same time and space, which means that mmWave radar and vision information need to be~calibrated.
	
	\item Information fusion: Object detection algorithms that fuse the sensing information of different sensors to achieve the optimal performance are~essential.
	
\end{itemize}

In order to solve the above challenges, first, it is essential to analyze the characteristics of different sensors and select the appropriate sensor. Second, it is necessary to study the sensor calibration, which involves the coordinate transformation between different sensors, the~filtering of invalid sensing information, and~the error calibration. Last but not least, it is necessary to study the sensing information fusion and realize the improvement of sensing capability through the complement of different sensors, which involve the mmWave radar and vision fusion~schemes.

In recent years, there have been some surveys on object detection, but~most of them focus on visual solutions. Ref.~\cite{019} summarized the vision-based object detection solutions and~conducted a~comprehensive and detailed survey for object detection algorithms before and after the advent of deep learning. It is helpful for understanding vision-based object detection, but~its detection targets were not only vehicles and there is no relevant content about fusion of radar and vision. Ref.~\cite{020} summarized the functions of each sensor module in~the~advanced driver assistance system (ADAS), but~lacks the explanation of the fusion between radar and vision. Ref.~\cite{021} summarized the multi-sensor fusion and vehicle communication technology for autonomous driving, involving the fusion of cameras, mmWave radar, lidar, global positioning system (GPS), and other sensors and vehicle to everything (V2X) technology. It focused on the classification of sensor fusion schemes, but~lacks sufficient and in-depth review on the radar and vision fusion. In~this article, we focus on the low-cost mmWave radar and vision fusion solutions. The~process of mmWave radar and vision fusion consists of sensor selection, sensor calibration, and sensor fusion, each of which is reviewed in detail. Specifically, we classified the fusion methods into three-level fusion schemes including data level, decision level, and feature level schemes.

\begin{figure}[H]
	\includegraphics[width=0.7\textwidth]{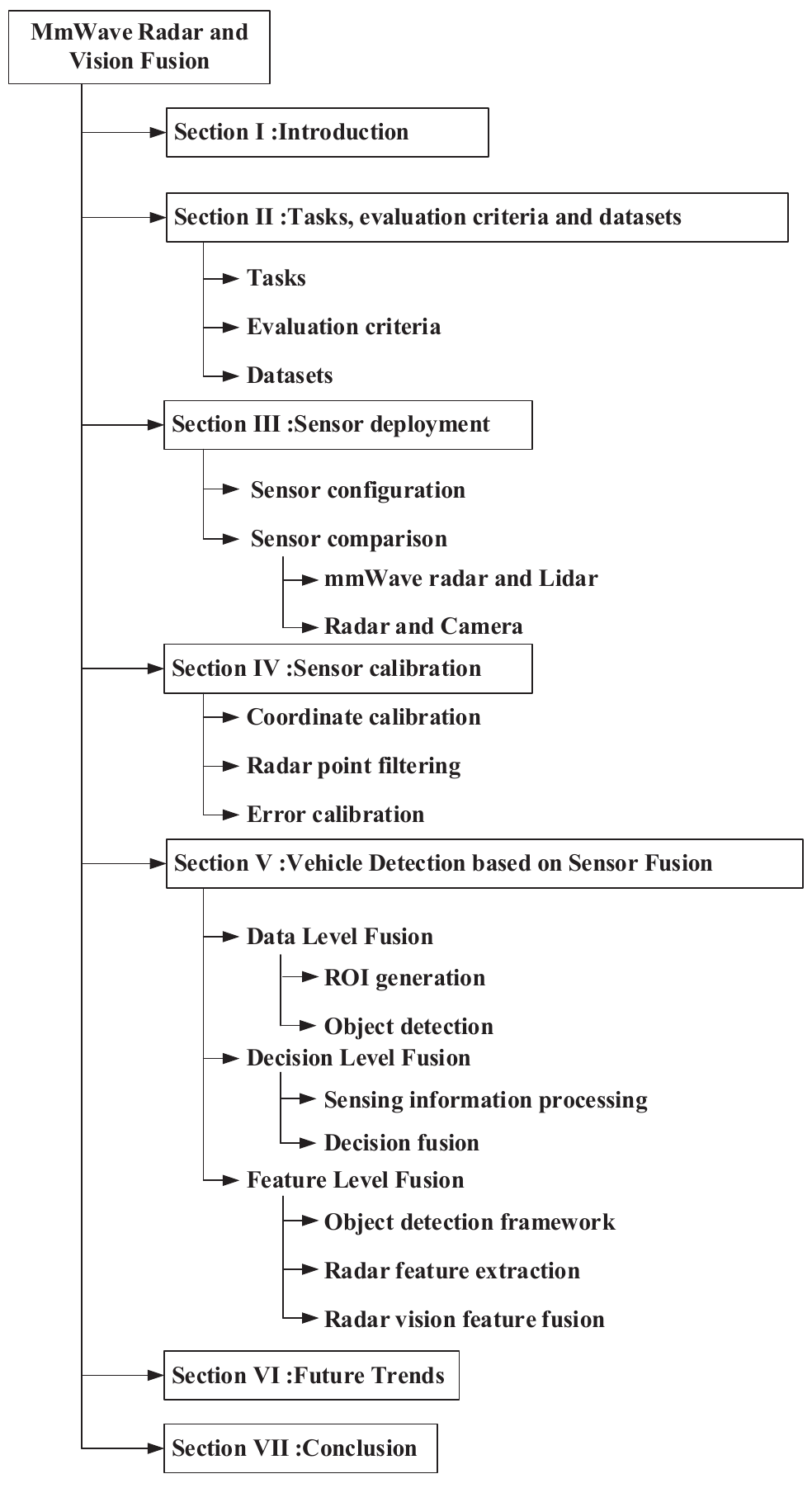}
	\caption{The organization of this~article.}
	\label{The organization of this paper}
\end{figure}

As shown in~Figure \ref{The organization of this paper},
the rest of this article is organized as follows.
{Section} \ref{sec2} introduces the tasks of object detection, evaluation criteria, and the public dataset.
Section~\ref{sec3} reviews the sensor configuration schemes of some automobile manufacturers and the comparison of the sensors.
Section~\ref{sec4} reviews the sensors calibration technologies.
Section~\ref{sec5} reviews sensor fusion schemes.
Section~\ref{sec6} analyzes the development trends of fusion of radar and vision for autonomous driving.
Finally, we summarize this article in~Section \ref{sec7}.

\unskip

\section{Tasks, Evaluation Criteria, and~Datasets}\label{sec2}

\subsection{Tasks}
Object detection is of vital importance in~the~field of autonomous driving. In~addition to target classification, the~detection tasks include positioning of the existing objects in~the~input~image.

The goal of two-dimensional (2D) object detection is to select the detected obstacle targets in~the~vision image of the vehicle with a 2D bounding box. The~targets are then classified and positioned. The~positioning here refers to the positioning of the targets in~the~image rather than the positioning of targets relative to the vehicle in~the~real~world.

In three-dimensional (3D) object detection, the~processes include the selection of the detected obstacle targets in~the~vision image of the vehicle with a 3D bounding box and the~classification and positioning of the targets. The~positioning is not only the positioning of the selected target in~the~image, but~also determining the posture and position of the objects in~the~real~world.

\subsection{Evaluation~Criteria}
Average precision (AP) and average recall (AR), representing accuracy and regression in~object detection, are commonly used as evaluation indices. The~precision-recall (PR) curve can be obtained by taking recall values and precision values as the horizontal and vertical axes, respectively. The~mean average precision (mAP), which represents the consolidated results of the detection model, can be obtained by calculating the average AP value of all~classes.

Take the KITTI dataset, which is widely used in~autonomous driving, as~an example. For~2D object detection, the~correctness of object positioning is determined by comparing whether the intersection over union (IoU) between the detection bounding box and the annotated bounding box is greater than the threshold \cite{025}. It uses the mAP model proposed in~\cite{022} for performance evaluation. The~calculation method is to set 11 equally spaced thresholds in~[0, 1]. For~recall value greater than each threshold, there is a~maximum precision value, and~the average value of 11 precise values is the final mAP~value.

However, 3D object detection currently is more attractive in~the~studies of autonomous driving. The~KITTI officially stipulated that for vehicles, correct prediction requires that the predicted 3D box overlaps the real 3D box by more than 70\%, whereas for pedestrians and bicycles, 50\% overlap of the 3D bounding box is required~\cite{093}. In~addition, KITTI adopted the recommendations of the Mapillary team in~\cite{023}, which proposed that 40 recall positions could calculate the mAP more accurately than the 11 recall positions~method.

\subsection{Datasets}
Datasets are very important for the research of object detection. This section will briefly introduce some representative datasets for autonomous driving. Table~\ref{tab1} shows the basic features of some widely used~datasets.

\begin{table}[H]
	\caption{{Autonomous} driving datasets. 
		``Y'' indicates the existence of the sensing information in this dataset and ``N'' indicates the absence of the sensing information in this dataset.}
	\label{tab1}
	\renewcommand{\arraystretch}{1.1} 
	\begin{center}
		\begin{tabular}{|p{100pt}<{\centering}|p{30pt}<{\centering}|p{35pt}<{\centering}|p{25pt}<{\centering}|p{25pt}<{\centering}|}
			\hline
			Dataset & Release Year & RGB image & Radar & Lidar \\
			\hline
			Apolloscape & 2018 & Y & N & Y \\
			\hline
			KITTI & 2012 & Y & N & Y \\
			\hline
			Cityscapes & 2016 & Y & N & N \\
			\hline
			Waymo Open Dataset & 2019 & Y & N & Y \\
			\hline
			nuScense & 2019 & Y & Y & Y \\
			\hline
		\end{tabular}
	\end{center}
	
\end{table}
\unskip

\subsubsection{{Apolloscape}}  
ApolloScape~\cite{024} is part of the Apollo Open Platform created by {Baidu} in~2017. It uses a~Reigl lidar to collect point cloud. The~3D point cloud generated by Reigl is more accurate and denser than the point cloud generated by {Velodyne}. Currently, ApolloScape has opened 147,000 frames of pixel-level semantically annotated images, including perceptual classification and road network data, etc. These high-resolution images with pixel-by-pixel semantic segmentation and annotation come from a~10 km distance measurement across three cities. Moreover, each area was repeatedly scanned under different weather and lighting~conditions. 

\subsubsection{{KITTI}}
The KITTI dataset~\cite{025}, established by the {Karlsruhe Institute of Technology} in~Germany and the {Toyota American Institute of Technology} in~the USA, is currently the most commonly used autonomous driving dataset. The~team used a~Volkswagen car equipped with cameras and Velodyne lidar to drive around Karlsruhe, Germany for 6 hours to record the traffic information. The~dataset provides the raw image and accurate 3D bounding box with class label for every sequence. The~object classes include cars, vans, trucks, pedestrians, cyclists, and~trams. 

\subsubsection{{Cityscapes}}
The Cityscapes dataset~\cite{026} is jointly provided by three German labs: {Daimler}, {Max planck institute informatik}, and {Technische Universit\"{a}t Darmstadt}. It is a~semantic understanding image dataset of urban street scenes. It primarily contains 5000 high-quality pixel-level annotated images of driving scenes in~an urban environment (2975 for training, 500 for validation, 1525 for test, for 19 categories in~total) from over 50 cities. In~addition, it has 20,000 rough-annotated~images.

\subsubsection{Waymo Open {Dataset}}
The Waymo dataset~\cite{027} is an open source project of {Waymo}, an~autonomous driving company under {Alphabet Inc}. It consists of labeled data collected by Waymo self-driving cars under various conditions, including more than 10 million miles of autonomous driving mileage data covering 25 cities. The~dataset includes lidar point clouds and vision images. Vehicles, pedestrians, cyclists, and~signs have been meticulously marked. The~team captured more than 12 million 3D annotations and 1.2 million 2D~annotations.

\subsubsection{{nuScenes}}
The nuScenes dataset~\cite{028}, established by {nuTonomy}, is the largest existing autonomous driving dataset. It is the first dataset that is equipped with the full autonomous vehicle sensors. This dataset not only provides camera and lidar data, but~also contains radar data, and it is currently the only dataset with radar data. Specifically, the~3D bounding box annotation provided by nuScenes not only contains 23 classes, but~also has 8 attributes including pedestrian pose, vehicle state, etc.

\section{Sensor~Deployment}\label{sec3}

This section introduces the sensor deployment schemes for autonomous driving vehicles. Through the analysis of the sensor deployment for vehicles equipped with autonomous driving systems launched by several major automobile manufacturers, it can be found that mmWave radar, lidar, and~cameras are the main sensors for vehicles to detect the obstacles. In~the~following sections, the~sensor deployment schemes including sensor configuration, sensor comparison, and selection, are reviewed in~detail.

\subsection{Sensor~Configuration}

As shown in  Table~\ref{tab2}, the overwhelming majority of automobile manufacturers have adopted a~sensor configuration scheme combining radar and cameras. In~addition to Tesla, other manufacturers have used the fusion sensing technology combining lidar, mmWave radar, and cameras. It can be concluded that the sensing solution using the fusion of radar and vision is the current mainstream trend in~the~field of obstacle detection for autonomous driving vehicles. The~reason is that the radar and the camera have complementary characteristics. The~specific details will be explained in~Section \ref{sec3.2}.

\begin{table}[H]
	\caption{Autonomous driving sensor solutions of some {manufacturers}~\cite{029,030,031,032,033,034}.} 
	\label{tab2}
	\renewcommand{\arraystretch}{1.1} 
	\begin{center}
		\begin{tabular}{|m{0.15\textwidth}<{\centering}|m{0.15\textwidth}<{\centering}|m{0.15\textwidth}<{\centering}|m{0.15\textwidth}<{\centering}|}
			\hline
			Company & Autonomous driving system & Sensor configuration & SAE autonomous driving level \\
			\hline
			Tesla & Autopilot & 8 cameras, 12 ultrasonic radars mmWave radar & Level 2 \\
			\hline
			Baidu &	Apollo & Lidar mmWave radar Camera & Level 4 \\
			\hline
			NIO	& Aquila & Lidar 11 cameras 5 mmWave radars 12 ultrasonic radars &	Level 3 \\
			\hline
			Xpeng & XPILOT & 6 cameras 2 mmWave radars 12 ultrasonic radars & Level 3 \\
			\hline
			Audi & Traffic Jam Pilot & 6 cameras 5 mmWave radars 12 ultrasonic radars Lidar & Level 3 \\
			\hline
			Mercedes Benz & Drive Pilot & 4 panoramic cameras Lidar mmWave radar & Level 3 \\
			\hline
			
		\end{tabular}
	\end{center}
\end{table}
\unskip

\subsection{Sensor~Comparison}\label{sec3.2}
Ref.~\cite{038} investigated the advantages and disadvantages of mmWave radar, lidar, and~cameras in~various applications. This article compares the characteristics of the three sensors, as~shown in~Table \ref{tab3}. According to Table~\ref{tab3}, three sensors have complementary~advantages.

\begin{table}[H]
	
	\caption{Comparison of mmWave radar, lidar, and {camera}~\cite{035,036,037,038,039}. 
		{``1''--``6'' denote the levels from ``extremely low'' to ``extremely high''}.}
	\label{tab3}
	
	\renewcommand{\arraystretch}{1.1} 
	\begin{center}
		\begin{tabular}{|m{0.15\textwidth}<{\centering}|m{0.15\textwidth}<{\centering}|m{0.15\textwidth}<{\centering}|m{0.15\textwidth}<{\centering}|}
			\hline
			Sensor type & mmWave radar & Lidar & Camera \\
			\hline
			Range resolution & {\color{cyan!75!black}4} & {\color{cyan!75!black}5} & {\color{cyan!75!black}2} \\
			\hline
			Angle resolution & {\color{cyan!75!black}4} & {\color{cyan!75!black}5} & {\color{cyan!75!black}6} \\
			\hline
			Speed detection & {\color{cyan!75!black}5} & {\color{cyan!75!black}4} & {\color{cyan!75!black}3} \\
			\hline
			Detection accuracy & {\color{cyan!75!black}2} & {\color{cyan!75!black}5} & {\color{cyan!75!black}6} \\
			\hline
			Anti-interference performance & {\color{cyan!75!black}5} & {\color{cyan!75!black}5} & {\color{cyan!75!black}6} \\
			\hline
			Requirements for weather conditions & {\color{cyan!75!black}1} & {\color{cyan!75!black}4} & {\color{cyan!75!black}4} \\
			\hline
			Operating hours & All weather & All weather & Depends on light conditions \\
			\hline
			Cost and processing overhead & {\color{cyan!75!black}2} & {\color{cyan!75!black}4} & {\color{cyan!75!black}3} \\
			\hline
		\end{tabular}
	\end{center}
	
\end{table}
\unskip
\subsubsection{mmWave Radar and~Lidar}
As a~common and necessary sensor on autonomous driving vehicles, mmWave radar has the characteristics of long range detection, low cost, and detectability of dynamic targets. Due to these advantages, the~vehicle’s sensing ability and safety have been improved~\cite{035}. Compared with lidar, the~advantages of mmWave radar are mainly reflected in~the~aspects of coping with severe weather and low deployment cost~\cite{038}. In~addition, it has the following~advantages.
\begin{itemize}
	
	\item The~mmWave radar can detect obstacles within 250 m, which is of vital importance to the security of autonomous driving, whereas the detection range of lidar is within 150~m~\cite{040}. 
	
	\item The~mmWave radar can measure the relative velocity of the target vehicle based on the Doppler effect with the resolution of 0.1m/s, which is critical for vehicle decision-making in~autonomous driving~\cite{040}.
	
\end{itemize}

{Compared with mmWave radar, lidar has the following advantages~\cite{036,037}. }
\begin{itemize}
	
	{\item Lidar has relatively higher angle resolution and detection accuracy than mmWave radar. Additionally, the~mmWave radar data is~sparser.
		
		\item The measurements of lidar contain semantic information and satisfy the perception requirements of advanced autonomous driving, which mmWave radar~lacks.
		
		\item The~clutter cannot be completely filtered out from mmWave radar measurements, leading to errors in~radar signal processing.}
	
\end{itemize}

Lidar applies laser beams to complete real-time dynamic measurement, establishing a~3D environment model, which supports the prediction of the surrounding environment and~the position and velocity of targets. Due to the characteristics of the laser, its propagation is less affected by clutter, and~the ranging is non-coherent and large-scale. According to~\cite{039}, the~detection range of lidar can reach more than 50 m in~the~longitudinal direction. The~detection circle radius in~the~periphery of the vehicle reaches more than 20 m. The~rear end of the vehicle reaches a~detection range of more than 20 m, with~high resolution of angular and range. However, the~field of view (FOV) of lidar is~limited.

The price is one of the factors limiting the massive application of lidar. However, with~the development of lidar technology, its cost is showing a~downward trend. The~more important factor is that lidar is highly affected by bad weather conditions.{ When encountering heavy fog, rain, and snow, the~attenuation of the laser is greatly increased, and~the propagation distance is greatly affected, thereby reducing its performance~\cite{038}.}

\subsubsection{Radar and~Camera}
Radar is the best sensor for detecting distance and radial speed. It has ``all-weather'' capability, especially considering that it can still work normally at night. However,~radar cannot distinguish color and has poor capability for target classification~\cite{038}. Cameras have good color perception and classification capabilities, and~the angle resolution capability is not weaker than that of lidar~\cite{038}. However, they are limited in~estimating speed and distance~\cite{039}. In~addition, image processing relies on the powerful computing power of the vehicular chip, whereas the information processing of mmWave radar is not required. Making full use of radar sensing information can greatly save computing resources~\cite{038}.

Comparing the characteristics of radar and cameras, it can be found that there are many complementary features between them. Therefore, the~application of radar and vision fusion perception technology in~the~field of obstacle detection can effectively improve the perception accuracy and enhance the object detection capability of autonomous vehicles. {Either mmWave radar or lidar and vision fusion are helpful. The~advantages of the two fusion schemes come from the respective advantages of mmWave radar and lidar. In~future research, the~fusion of mmWave radar, lidar, and vision may have greater potential.}

\section{Sensor~Calibration}\label{sec4}

\begin{figure}[H]
	\includegraphics[width=0.9\textwidth]{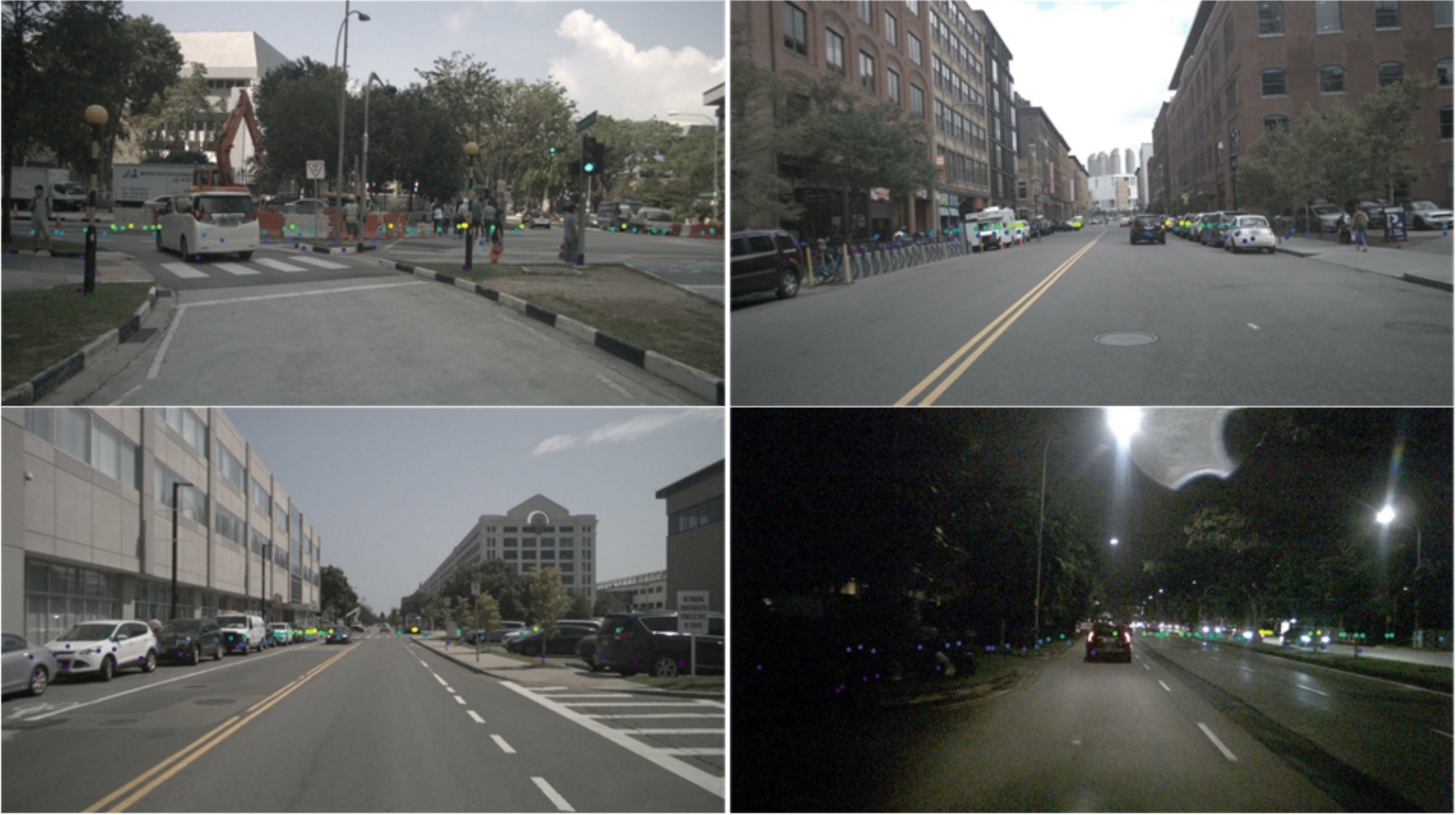}
	\caption{{Radar points} are projected on the~image and rendered into different colors.}
	\label{Radar points are projected on the image}
\end{figure}

Due to the difference in~the~spatial location and sampling frequency of different sensors, the~sensing information of different sensors for the same target may not match. Hence, calibrating the sensing information of different sensors is necessary. The~detection information returned by mmWave radar is radar points, and cameras receive visual images.{ We selected the camera and mmWave radar data from nuScenes~\cite{028} as an example. The~data provided by this dataset have been processed by frame synchronization, so time synchronization is not required, and~Figure~\ref{Radar points are projected on the image} can be obtained through spatial coordinate transformation. The~RGB value of the radar point is converted from the three physical quantities of transverse velocity, longitudinal velocity, and distance, and~the color of the radar point represents the physical state of the object corresponding to the radar point.} Generally speaking, sensor calibration involves coordinate calibration~\cite{041,049,051,061,062,077,079}, radar point filtering~\cite{049,051}, and error calibration~\cite{076,085,087}.

\subsection{Coordinate~Calibration}
The purpose of coordinate calibration is to match the radar points to the objects in~the~images. 
For coordinate calibration, the~most common methods are classified into coordinate transformation method~\cite{049,077}, sensor verification method~\cite{041,062,079}, and vision based method~\cite{050,051}, which are reviewed as~follows.

\begin{itemize}
	
	\item Coordinate transformation method: The~coordinate transformation method unifies the radar information and vision information under the same coordinate system through matrix operations. In~\cite{077}, space calibration was completed by the method of coordinate transformation according to the spatial position coordinates of mmWave radar and vision sensors. For~the time inconsistency caused by different sensor sampling rates, the~thread synchronization method is adopted to realize the acquisition of the image frame and mmWave radar data simultaneously. Ref.~\cite{049} used the point alignment method based on pseudo-inverse, which ~obtains the coordinate transformation matrix by using the least square method. The~traditional coordinate transformation cannot generate the accurate position of the target, which brings errors to the final results. In~\cite{048}, Wang~et~al. proposed a~calibration experiment to project the real coordinates into the radar detection map without special tools and radar reflection intensity, which weakens the dependence on calibration~errors.
	
	\item Sensor verification method: The~sensor verification method calibrates multiple sensors to each other with the detection information of different sensors on the same object. In~\cite{041}, the~sensor verification consists of two steps. First, the~target list is generated by radar, and~then the list is verified by the vision information. In~\cite{079}, after~the coordinate transformation of radar, the~image is first searched roughly and then compared with the radar information. The~result of the comparison divides the targets into two types: matched target and unmatched target. In~\cite{062}, Streubel~et~al. designed a~fusion time slot to match the objects detected by radar and vision in~the~same time~slot. 
	
	\item Vision based method: In~\cite{050}, the~motion stereo technology was used to achieve the matching of radar objects and image objects. In~\cite{051}, Huang~et~al. used adaptive background subtraction to detect moving targets in~the~image, generate candidate areas, and~verify the targets by judging whether the radar points are located in~the~candidate~areas. 
	
\end{itemize}

\subsection{Radar Point~Filtering}

The purpose of radar point filtering is to filter out noise and useless detection results to avoid misjudgments caused by these radar points. In~\cite{049}, Guo~et~al. proposed a~method for noise filtering and effective target extraction using intra-frame clustering and inter-frame tracking information. In~\cite{051}, the~radar points were filtered by the speed and angular velocity information obtained by mmWave radar. The~invalid radar points were then filtered, which reduces the impact of stationary targets such as trees and bridges on mmWave~radar.

{At present, in~the~field of autonomous driving, researchers mostly use open source datasets for training, and~the objects they want to detect are generally vehicles or pedestrians. Therefore, the method of~\cite{051} can be applied to filter the radar points according to the speed and other information to exclude non-vehicle and non-pedestrian objects.}

\subsection{Error~Calibration}

Due to errors in~sensors or mathematical calculations, there may be errors in~the~calibrated radar points. Some articles have proposed methods to correct these errors. In~\cite{076}, a~method based on interactive fine-tuning was proposed to make ultimate rectification to the radar points projected on the vision image. The authors in~\cite{087} proposed an improved extended Kalman filter (EKF) algorithm to model the measurement errors of different sensors. In~\cite{085}, the~influence of various coordinates on detection results was analyzed and discussed, and~a~semi-integral Cartesian coordinate representation method was proposed to convert all information into a~coordinate system that moves with the host~vehicle.

{With the current use of open source datasets, error calibration is not required. However, if~the dataset is self-made, radar filtering and error correction are necessary technical~steps.}

\section{Vehicle Detection Based on Sensor~Fusion}\label{sec5}

Generally speaking, there are three fusion levels for mmWave radar and vision, including data level, decision level, and feature level. Data level fusion is the fusion of data detected by mmWave radar and cameras, which has the minimum data loss and the highest reliability. Decision level fusion is the fusion of the detection results of mmWave radar and cameras. Feature level fusion requires extracting radar feature information and then fusing it with image features. The~comparison of the three fusion levels is provided in~Table \ref{tab4}.

\begin{table}[H]
	\caption{Summary of the three fusion~levels.}\label{tab4}
	\renewcommand{\arraystretch}{1.1} 
\begin{center}
	\begin{tabular}{|m{0.15\textwidth}<{\centering}|m{0.25\textwidth}|m{0.25\textwidth}|}
		\hline
		Fusion level & \makecell[c]{Advantages} & \makecell[c]{Disadvantages} \\
		\hline
		Data level  & Minimum data loss and the highest reliability & Depending on the number of radar points\\
		\hline
		Decision level  & Making full use of sensor information & Modeling the joint probability density function of sensors is difficult \\
		\hline
		Feature level  & Making full use of feature information and achieving best detection performance & Computing is complicated, the overhead of radar information transformation \\
		\hline
	\end{tabular}
\end{center}
	
\end{table}

\subsection{Data Level~Fusion}
\begin{figure}[H]
	\includegraphics[width=0.9\textwidth]{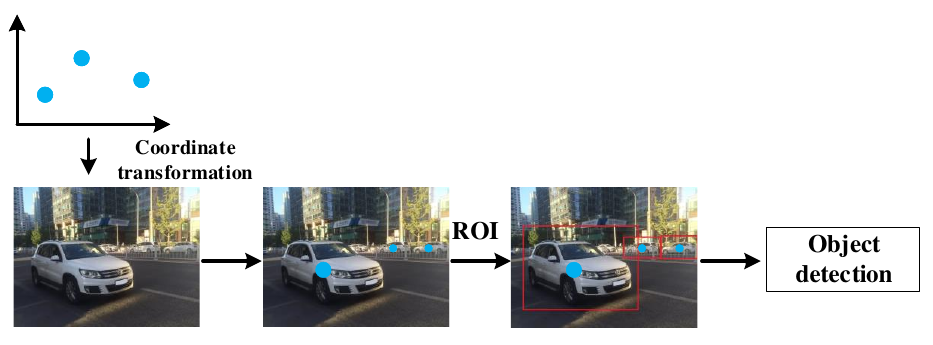}
	\caption{Data level fusion.}
	\label{Data level fusion}
\end{figure}

{Data level fusion was a mature fusion scheme and has not been the mainstream research trend at present. However, its idea of fusing different sensor information is still useful for reference.} As shown in~Table \ref{tab5}, data level fusion first generates the region of interest (ROI) based on radar points~\cite{041,049,071,075}. The~corresponding region of the vision image is then extracted according to the ROI. Finally, feature extractor and classifier are used to perform object detection on these images~\cite{042,045,048,049,070,072,074,075,078,079}. Some literatures use neural networks for object detection and classification~\cite{046,072}. For~data level fusion, the~number of effective radar points has a~direct impact on the final detection results. If~there is no radar point in~a~certain part of the image, this part will be ignored. This scheme narrows the searching space for object detection, saves computational resources, and~leaves a~hidden security danger at the same~time. The data level fusion process is shown in~Figure~\ref{Data level fusion}.

\begin{table}[H]
	\caption{{Summary} of data level fusion.}\label{tab5}
	
\renewcommand{\arraystretch}{1.1} 
\begin{center}
	\begin{tabular}{|c|c|m{0.12\textwidth}<{\centering}|m{0.35\textwidth}|}
		\hline
		\multicolumn{2}{|c|}{} & Reference & \makecell[c]{Contribution} \\
		\hline
		\multicolumn{2}{|c|}{\multirow{3}{*}{ROI generation}} &\cite{041} &Using radar points to increase the speed of ROI generation. \\
		\cline{3-4}
		\multicolumn{2}{|c|}{} &\cite{049}&{\color{cyan!75!black}Propose the conclusion that the distance determines the initial size of the ROI.} \\
		\cline{3-4}
		\multicolumn{2}{|c|}{} &\cite{071}&{\color{cyan!75!black}Extending ROI application to overtaking detection.} \\
		\hline
		\multirow{7}{*}{{Object detection }} & \multirow{2}{*}{{Image preprocessing }} & \cite{045} \cite{049} \cite{072} &Using histogram equalization, grayscale variance and contrast normalization to preprocess the image. \\
		\cline{3-4}
		&    & \cite{048} \cite{072} \cite{074}&{\color{cyan!75!black}Image segmentation preprocessing with radar point as reference center.}  \\
		\cline{2-4}
		
		& \multirow{2}{*}{{Feature extraction }}  & \cite{042} \cite{043} \cite{086} \cite{074} \cite{075} \cite{078}  &Using features such as symmetry and shadow to extract vehicle contours.\\
		\cline{3-4}
		&   & \cite{044} \cite{045}&Using Haar-like model for feature extraction.  \\
		\cline{2-4}
		
		& \multirow{3}{*}{{Object classification }} & \cite{045} & Adaboost algorithm for object classification. \\
		\cline{3-4}
		&   & \cite{070} \cite{079} & SVM for object classification.  \\
		\cline{3-4}
		&   & \cite{046} \cite{072} & Neural network-based classifier for object classification.\\
		\hline   
	\end{tabular}
\end{center}
	
\end{table}
\unskip

\begin{table}[H]
	\caption{{Summary} of decision level fusion.}\label{tab6}
	\renewcommand{\arraystretch}{1.1} 
	\begin{center}
	
			\begin{tabular}{|c|m{0.2\textwidth}<{\centering}|m{0.12\textwidth}<{\centering}|m{0.3\textwidth}|}
				\hline
				\multicolumn{2}{|c|}{} & Reference & \makecell[c]{Contribution} \\
				\hline
				\multirow{5}{*}{{\makecell[c]{Sensing information\\  processing} }} &
				\multirow{1}{*}{{Radar information }} & \cite{052}\cite{058} & The techniques involved in radar signal processing and what physical states can be obtained from radar information are analyzed. \\
				\cline{2-4}
				
				& \multirow{4}{*}{{\makecell[c]{Image \\object detection} }} & \cite{058} & Pedestrian detection using feature extraction combined with classifiers. \\
				\cline{3-4}
				&   &  \cite{063}&Detecting objects in depth images with MeanShift algorithm.  \\
				\cline{3-4}
				&   &  \cite{064}&An upgraded version of \cite{063}, using MaskRCNN for target detection.  \\
				\cline{3-4}
				&   &  \cite{088}\cite{089}\cite{090}&Using one-stage object detection algorithm YOLO for radar vision fusion object detection tasks.  \\
				\hline
				\multirow{8}{*}{{Decision fusion }} & \multirow{2}{*}{{\makecell[c]{Based on \\Bayesian theory} }} & \cite{053} &Proposing Bayesian programming to solve multi-sensor data fusion problems through probabilistic reasoning \\
				\cline{3-4}
				&    &  \cite{054}&A dynamic fusion method based on Bayesian network is proposed to facilitate the addition of new sensors.  \\
				\cline{2-4}
				
				& \multirow{3}{*}{{\makecell[c]{Based on\\ Kalman filter} }}  & \cite{055}  &Proposing a decision level fusion filter based on EKF framework.\\
				\cline{3-4}
				&   &  \cite{059}&  The proposed fusion methon can track the object simultaneously in 3D space and 2D image plane.\\
				\cline{3-4}
				&   &  \cite{060}&Functional equivalence of centralized and decentralized information fusion schemes is demonstrated.  \\
				\cline{2-4}
				
				& \multirow{1}{*}{{\makecell[c]{Based on Dempster\\ Shafer theory} }} & \cite{058} & A decision level sensor fusion method based on Dempster-Shafer is proposed. \\
				\cline{2-4}
				
				& \multirow{2}{*}{{\makecell[c]{based on \\Radar validation} }} & \cite{056} &Using radar detection results to validate visual's. \\
				\cline{3-4}
				&    &  \cite{057}&Using radar information to correct vehicle position information in real time to achieve object tracking.  \\
				\hline   
		\end{tabular}
	\end{center}
\end{table}
\vspace{-9pt}

\begin{table}[H]
	\caption{{Summary} of feature level fusion.} \label{tab7}
	
	\renewcommand{\arraystretch}{1.1} 
	\begin{center}
		
			\begin{tabular}{|c|m{0.12\textwidth}<{\centering}|m{0.4\textwidth}|}
				\hline
				\multicolumn{1}{|c|}{} & Reference & \makecell[c]{Technology Features} \\
				\hline
				
				\multicolumn{1}{|c|}{\multirow{3}{*}{Fusion framework}} &\cite{065} &Based on SSD framework improvement, concatenation fusion is used. \\
				\cline{2-3}
				\multicolumn{1}{|c|}{} &\cite{066}&A fusion framework similar to YOLO structure is proposed named RVNet. \\
				\cline{2-3}
				\multicolumn{1}{|c|}{} &\cite{067}&Proposing CRF-Net built on the VGG backbone network and RetinaNet, and the radar input branch is extended. \\
				\cline{2-3}
				\multicolumn{1}{|c|}{} &\cite{069}&Join the radar branch based on the FCOS detection framework and  embedded SAF module. \\
				
				\hline
				
				\multicolumn{1}{|c|}{\multirow{3}{*}{Radar feature extraction}} &\cite{068} &Proposing a network named CMGGAN which can generate environmental images. \\
				\cline{2-3}
				\multicolumn{1}{|c|}{} &\cite{066}\cite{069}&Using a new radar feature description method called radar sparse image, the detected objects are presented as radar points. \\
				\cline{2-3}
				\multicolumn{1}{|c|}{} &\cite{067}&Stretching the radar points in the radar sparse image vertically to supplement the height information. \\
				
				\hline
				
				\multicolumn{1}{|c|}{\multirow{3}{*}{Fearure fusion}} &\cite{065}\cite{066}\cite{067} &The fusion method of concatenation and element-wise addition is adopted. \\
				\cline{2-3}
				\multicolumn{1}{|c|}{} &\cite{069}&A feature fusion block named spatial attention fusion is proposed which uses attention mechanism. \\
				\hline
		\end{tabular}
	\end{center}
	
\end{table}
\unskip

\subsubsection{ROI~Generation}
ROI is a~selected area in~the~image, which is the focus of target detection. Compared with a~pure image processing scheme, the~data-level fusion scheme uses radar points to generate ROI, which can significantly improve the speed of ROI generation~\cite{041}. The~size of the initial ROI is determined by the distance between the obstacle and mmWave radar~\cite{049}. The~improved front-facing vehicle detection system proposed by~\cite{071} can detect overtaking with a~high detection rate. This method focuses on the area where overtaking is about to occur. The~overtaking is determined by checking two specific characteristics of vehicle speed and movement angle. In~\cite{075}, a~square with 3 meters of side length centered on the radar point is taken as the~ROI.
\subsubsection{Object~Detection}
Because of the uncertainty of the position and size of the objects in~the~image, the~object detection based on vision often adopts sliding window and multi-scale strategy, which produces a~fair amount of candidate boxes, resulting in~low detection efficiency. The mmWave radar and vision fusion scheme can avoid the sliding window method, which reduces the computational cost and improves the detection efficiency. Object detection tasks focus on the vision image processing, and the~task can be divided into three steps: image \mbox{preprocessing~\cite{045,048,049,072,074}}, image feature extraction~\cite{042,043,044,045,061,072,074,075,078}, and object \mbox{classification~\cite{045,070,072,079}}. 

\begin{itemize}
	\item Image Preprocessing 
\end{itemize}

In order to remove the noise in~the~image and enhance the feature information, so as to facilitate the subsequent feature extraction and object classification tasks, image preprocessing is necessary. The~methods of image preprocessing mainly include histogram equalization, gray variance, contrast normalization, and image~segmentation.

In~\cite{045}, Bombini~et~al. conducted a~series of tests on histogram equalization, grayscale variance, and contrast normalization in~order to obtain the invariance under different illumination conditions and cameras. They concluded that contrast normalization achieved better performance. In~\cite{049}, the~gradient histogram method was used to preprocess the image, and~an improved position estimation algorithm based on ROI was proposed, which can obtain a~smaller potential object region and further improve the detection efficiency. In~\cite{072}, median filtering, histogram equalization, wavelet transform, and Canny operator were used in~image~preprocessing.

In~\cite{048,072,074}, radar points were taken as the reference center to segment the image, and~then the object boundary was determined to improve the object detection~speed. 

\begin{itemize}
	\item Feature Extraction
\end{itemize}

The purpose of feature extraction is to transform the original image features into a~group of features with obvious physical or statistical significance, which is convenient for object detection. In~the~stage of image feature extraction, the~available vehicle features include symmetry and underbody shadow, etc.

In~\cite{042,043,078}, symmetry was used for ROI detection and~\cite{061} utilized shadow detection to obtain feature information, whereas in~\cite{086,075}, vertical symmetry and underbody shadow characteristics were comprehensively utilized for vehicle detection. In~\cite{075}, the~gradient information of the image was used to locate the boundary effectively, and~the method based on gradient vector flow (GVF) Snake was used to describe the accurate contour of the vehicle. As the color distribution of the object was stable during the movement, the~histogram matching method was feasible for tracking the~vehicle.

In~\cite{044}, Kadow~et~al. applied Haar-like model for feature extraction, which is a~classic feature extraction algorithm for face detection. In~order to improve detection rate, mutual information was used for Haar-like feature selection~\cite{045}.

In~\cite{074}, the~visual selective attention mechanism and prior information on visual consciousness during human driving was proposed to extract features and identify object contours from segmented images. The~shadow was extracted by histogram and binary image segment, and~the pedestrian edge was detected by 3 $\times$  3 image corrosion~template.

\begin{itemize}
	\item Object Classification
\end{itemize}

In the object classification stage, Adaboost, support vector machine (SVM), and other object classification algorithms are used to select the final box of the vehicle in~the~vision image. In~\cite{045}, the Adaboost algorithm is used to scan the ROI projected on the image plane. In~\cite{070,079}, SVM was used for object recognition and classification. Ref.~\cite{080} combined ROI image and Doppler spectrum information for object classification. Ref.~\cite{072} adopted object classification based on infrared image analysis. The~authors classified the objects into point objects and regional objects according to the object area, and~used a~neural network-based classifier to classify the regional targets~\cite{072}. Ref.~\cite{046} utilized multilayer in-place learning network (MILN) as a~classifier, which demonstrated superior accuracy in~two-class classification~tasks.

\subsection{Decision Level~Fusion}
\begin{figure}[H]
	\includegraphics[width=0.9\textwidth]{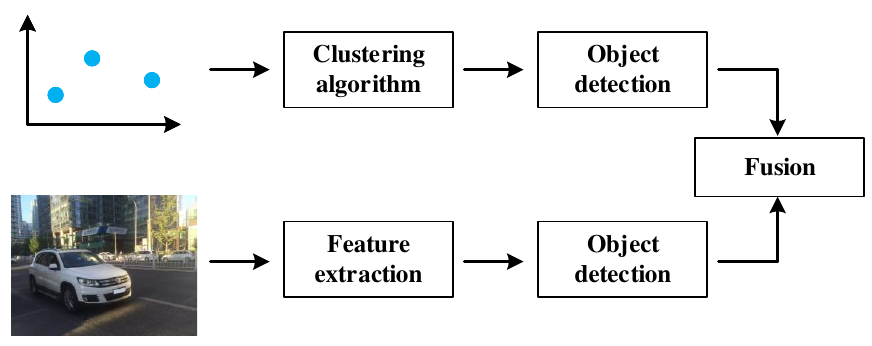}
	\caption{Decision level fusion.}
	\label{Decision level fusion}
\end{figure}

{Decision level fusion is the mainstream fusion scheme at present.}The process is shown in the table \ref{tab6}. The~advantage of radar is longitudinal distance, and the advantage of the vision sensor is horizontal field of view. Decision level fusion can take into account the advantages of both aspects and make full use of sensing information. The~challenge of the decision-level fusion filtering algorithm is to model the joint probability density function of the two kinds of detection information. This is due to the fact that the noise of the two kinds of detection information is different. The~decision level fusion mainly includes two steps: sensing information processing~\cite{052,058,063,064,088,090} and decision fusion~\cite{053,054,055,056,057,058,059,060,086}. The decision level fusion process is shown in~Figure~\ref{Decision level fusion}.

\subsubsection{Sensing Information~Processing}

The processing of sensing information includes radar information and vision information. Radar detection results generate a~list of objects and contain information such as the speed and distance of objects~\cite{052,058}. Vision information processing performs object detection algorithm on images, which includes traditional feature extraction combined with classifier~\cite{058,063} and convolutional neural network (CNN) \cite{064,088,090}.

\begin{itemize}
	\item Radar Information
\end{itemize}

A mmWave radar and vision fusion system was proposed for autonomous driving navigation and lane change in~\cite{052}, in~which the radar sensor obtains the target distance through fast Fourier transform (FFT), obtains the angular position of the target through digital wavefront reconstruction and beamforming, and~then analyzes the target position. In~\cite{058}, the~results of mmWave radar detection is a~list of possible objects in~the~radar FOV, and~each element in~the~list includes the distance, azimuth angle, and relative velocity of the detected~object.

\begin{itemize}
	\item  Image Object Detection
\end{itemize}

In the visual detection of~\cite{058}, a histogram was used to calculate edge information. Gradient histogram feature extraction and Boosting based classifier were then used to detect a pedestrian. Ref.~\cite{063} proposed an auxiliary navigation method combining mmWave radar and RGB depth sensor and~used the MeanShift algorithm to detect objects in~depth images. Furthermore, the~average depth of ROI determined the distance of the detected~object.

Ref.~\cite{064} has improved the vision data processing algorithm in~\cite{063}, and~applied mask R-CNN for object detection. In~\cite{088}, YOLO V2 is used for vehicle detection, and~the input is an RGB image of the size 224 $\times$ 224. Preprocessing subtracts the average RGB value of training set images from each pixel. Ref.~\cite{090} also applied YOLO V2 for vehicle detection and~proved that YOLO V2 is superior to faster R-CNN and SSD in~both speed and accuracy, which is more suitable for vehicle detection tasks. Ref.~\cite{089} applied YOLO V3 in~obstacle detection, and~its weight is pre-trained in~the COCO~dataset.

\subsubsection{Decision~Fusion}

The decision level fusion of vehicle detection fuses the detection results of different sensors. The~mainstream filtering algorithms apply Bayesian theory~\cite{053,054}, Kalman filtering framework~\cite{055,059,060}, and~Dempster Shafer theory~\cite{058}. In~some literatures, the~list of radar detection targets was used to verify vision detection results~\cite{056,057}. In~addition, ref.~\cite{086} proposed the motion stereo algorithm to adjust and refine the final detection~results.

\begin{itemize}
	\item Fusion Methods Based on Bayesian Theory
\end{itemize}

Ref.~\cite{053} proposed a~method based on Bayesian theory to solve the problem of multi-sensor data fusion by using a probabilistic reasoning method, which is called Bayesian programming. The~traditional multi-sensor fusion algorithms are no longer applicable when new sensors are added. The~fusion algorithm is modularized and generalized in~\cite{054}, and~a~dynamic fusion scheme based on a Bayesian network was proposed to improve the reusability of each fusion~algorithm.

\begin{itemize}
	\item Fusion Methods Based on Kalman Filter
\end{itemize}

Based on the EKF framework of the Lie group, a~decision level fusion filter using a~special Euclidean group is proposed in~\cite{055}. Ref.~\cite{059} proposed a~fusion framework that can track the detection object simultaneously in~3D space and a 2D image plane. An~uncertainty driven mechanism similar to a Kalman filter is used to equalize the sensing results of different qualities. In~\cite{060}, the~given image was first detected by radar to roughly search the target. A~trained spot detector was then used to obtain the object’s bounding box. The~information fusion method based on Kalman filter was adopted, and~the functional equivalence of centralized and decentralized information fusion schemes was~proved.
\newpage
\begin{itemize}
	\item Fusion Methods Based on Dempster Shafer Theory
\end{itemize}

Ref.~\cite{058} proposed decision level fusion based on the Dempster Shafer theory, taking the detection lists of multiple sensors as input, using one of them as a~temporary evidence grid and fusing it with the current evidence grid, and~finally performing clustering processing. The~target was identified in~the~evidence~grid.

\begin{itemize}
	\item Fusion Methods Based on Radar Validation
\end{itemize}

Ref.~\cite{056} overlapped the target list generated by vision detection and radar detection to generate a~unique vehicle list. The~radar data was used to verify the vision detection results. If~there was a~target matching the vision detection results in~the~radar data, a~blue box would be marked as a~strong hypothesis. Otherwise, if~there was no target, it would not be discarded: a~green box would be marked as a~weak hypothesis. Ref.~\cite{057} proposed a multi-target tracking (MTT) algorithm that can correct the tracked target list in~real time by evaluating the tracking score of the radar scattering center. The~stereo vision information is used to fit the contour of the target vehicle, and~the radar target matched with the target vehicle is used to correct its~position.

\subsection{Feature Level~Fusion}

\begin{figure}[H]
	\includegraphics[width=0.9\textwidth]{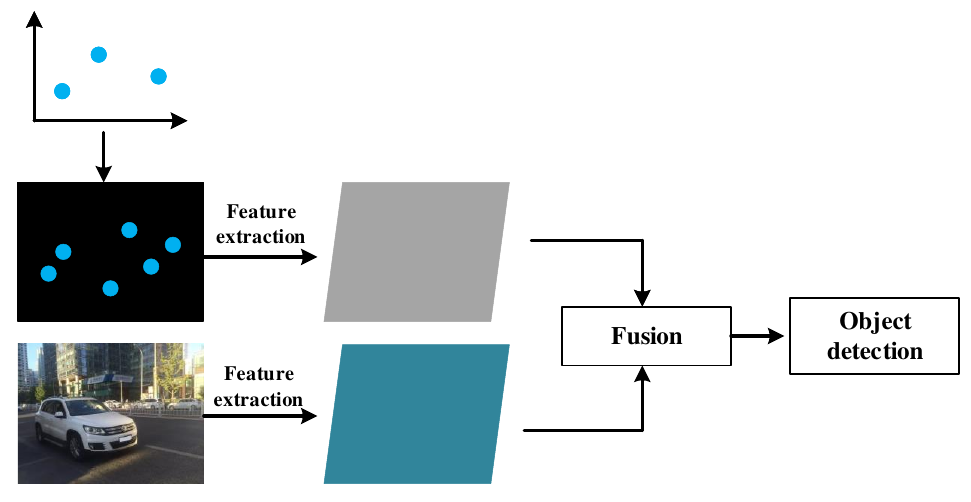}
	\caption{Feature level fusion.}
	\label{Feature level fusion}
\end{figure}

Feature level fusion is a~new scheme that has emerged in~recent years.The process is shown in the table \ref{tab7}. It is a~common approach to use an additional radar input branch in~feature level fusion methods~\cite{065,066,067,068,069}. The CNN-based objects detection model can effectively learn image feature information. By~transforming radar detection information into image form, the~detection model can learn radar and vision feature information simultaneously, and~the feature level fusion will be~realized. The feature level fusion process is shown in~Figure~\ref{Feature level fusion}.

\subsubsection{Object Detection~Framework}

Convolutional neural networks (CNNs) are widely applied in~object detection based on feature level fusion. At~present, some algorithms have achieved the results of superior performance, such as Faster RCNN, YOLO (V3), SSD, RetinaNet, etc. 

\begin{itemize}
	
	\item Detection Framework Based on CNN
\end{itemize}

Faster-RCNN~\cite{012} is a~widely used detector, and~its integration has been greatly improved compared with Fast-RCNN~\cite{011}. Its main contribution is the introduction of the regional advice network (RPN), which enables nearly costless area suggestion. The~RPN can simultaneously predict the target bounding box and categorization score for \mbox{each~location}.

YOLO~\cite{007} is the first one-stage detector, which is the abbreviation of ``You Only Look Once''. It adopts a~new detection idea: using a~single neural network to complete the detection task. It divides the vision image into multiple regions and predicts bounding box for each region. Therefore, the~detection speed has been remarkably improved. However, its positioning accuracy is reduced compared with the two-stage~detector. 

SSD~\cite{008} is another excellent one-stage detector. It eliminates the bounding box suggestion and pixel or feature resampling stage and~fundamentally improves the detection speed. The~detections in~different scales are generated from the feature maps of different scales, and~the detections are clearly separated by aspect ratio, such that the accuracy is significantly improved. As~a~one-stage detector with high detection speed, SSD also ensures the detection accuracy close to that of two-stage~detectors.

As to RetinaNet~\cite{009}, the~inventors discussed the reason why the accuracy of one-stage detector is lower than that of two-stage detector, and~concluded that this phenomenon is caused by the extreme foreground background class imbalance. They proposed a~new loss function named Focal loss to reshape the standard cross entropy loss, aiming to focus more attention on the objects that are difficult to classify during~training.

\begin{itemize}
	\item Fusion Framework Based on CNN
\end{itemize}

In~\cite{065}, feature level fusion was first applied in~mmWave radar and vision fusion. Its detection network was improved on the basis of SSD~\cite{008}. The~radar branch was concatenated after the second ResNet18 block of the vision~branch.

In~\cite{066}, a~new sensor fusion framework called RVNet was proposed, which was similar to YOLO~\cite{007}. The~input branches contain separate radar and vision branches and the output branches contain independent branches for small and large obstacles, respectively.

The CNN used in~\cite{067} was built on RetinaNet~\cite{009} with a~VGG backbone~\cite{091}, named CameraRadarFusionNet (CRF-Net). The~author expanded the network so that it can process an additional radar input branch. Its input branches were radar feature and vision feature, and~the output results are the 2D regression of the coordinates of the objects and the classification fraction of the~objects.

In~\cite{069}, the~author proposed a~new detection network called spatial attention fusion-based fully convolutional one-stage network (SAF-FCOS), which was built on FCOS~\cite{092}. The~radar branch was improved from ResNet-50, and~the vision image branch adopted a~two-stage operation block similar to ResNet-50. In~order to improve the detection accuracy, the~final object detection comprehensively utilized FCOS~\cite{092} and RetinaNet~\cite{009}.

\subsubsection{Radar Feature~Extraction}

The purpose of radar feature extraction is to transform radar information into image-like matrix information, because~radar information cannot be fused with image information directly. Radar feature extraction mostly adopts the method of converting radar points to the image plane to generate a radar image. The~transformed radar image with multiple channels contains all the environmental features detected by the radar. Each channel represents a~physical quantity such as distance, longitudinal speed, lateral speed, and so~on.

Ref.~\cite{068} proposed a~new Conditional Multi-Generator Generative Adversarial Network (CMGGAN), which takes the measured data of radar sensors to generate artificial, camera-like environmental images, including all environmental features detected by radar sensors. A~new description method for radar features was proposed in~\cite{066}, which is called radar sparse image. Radar sparse image is a~416  $\times$ 416 three-channel image, whose size directly corresponds to the size of the vision image. The~three channels contain radar point velocity and depth feature information. In~\cite{069}, Chang~et~al. converted the depth, horizontal, and vertical information at the radar point into real pixel values of different channels. For~the area without radar points, they set the pixel value to 0 and rendered the radar image with circles centered on radar points. Based on experimental results, the~anti-noise performance of the radar image saved in~the~PNG format is better than that of the jpg format. In~\cite{067}, considering the lack of height information in~radar detection results, Nobis~et~al. stretched the projected radar points in~the~vertical direction in~order to better integrate them with the image. The features of radar information were stored in~the~enhanced image in~the~form of pixel values. In~addition, a~ground truth noise filter was proposed to filter the invalid radar~points.

\subsubsection{Feature~Fusion}

The basic feature fusion methods can be classified into two kinds: concatenation and element-wise addition. The~former concatenates the radar feature matrix and the image feature matrix into a~multi-channel matrix, whereas the latter adds two matrices into~one.

In~\cite{065}, two fusion methods of concatenation and element-wise addition are set, and~the experimental results show that both fusion methods have improved the detection performance. The~element-wise addition method performs better on manually labeled test sets, and the concatenation method performs better on generated test sets. Refs.~\cite{066,067} both adopt the concatenation method. In~\cite{069}, a~new block named spatial attention fusion (SAF) was proposed for sensor feature fusion. An SAF block was used to generate an attention weight matrix to fuse radar and vision features. At~the same time,~\cite{069} compared the SAF method with three methods of element-wise addition, multiplication, and concatenation, and~the result shows that SAF has the best performance. In~addition,~\cite{069} conducted generalization experiments on faster R-CNN, and~the detection performance was also improved by the SAF~model.

\section{{Challenges and Future~Trends}}\label{sec6}
{\subsection{Challenges}
	
	For the object detection task, the~current research results have achieved superior performance, however, most of these results are 2D object detection. In~real autonomous driving scenarios, complex traffic environments often require 3D target detection to more accurately perceive environmental information. The~performance of the current 3D object detection network is far below the level of 2D detection. Therefore, improving the accuracy of 3D object detection is not only a~challenge in~the~field of autonomous driving, but~also a~major challenge in~the~task of object~detection.
	
	Challenges still exist for the fusion of mmWave radar and vision, which is the~focus of this article. The~biggest drawback of mmWave radar is the sparseness of radar features. Compared with vision images, mmWave radar provides very little information and cannot bring significant performance improvement. In~addition, whether the feature information of mmWave radar and vision can be further integrated and the associated mutual information between them has been mined remains to be studied. Therefore, mmWave radar vision fusion still has two major challenges: sparse perception information and more efficient fusion methods, which are also the two major challenges in~the~field of multi-sensor fusion.}

\subsection{Future~Trends}
For the future development of object detection in~the~field of autonomous driving, the authors of this article believe that there are {three} main trends. {One of them is 3D object detection. Improving the accuracy of 3D object detection will be a~major research trend. The~remaining two trends concern radar vision fusion.} On the one hand, it is necessary to integrate new sensing information, namely, adding new sensors, such as lidar, which has achieved superior performance in~autonomous driving; on the other hand, it is necessary to explore new ways of sensing information fusion, such as multimodal~fusion.

{\subsubsection{3D Object Detection in~Autonomous~Driving}
	
	Three-dimensional object detection can not only identify the position of the object in~the~image, but~can also detect the pose and other information of the object in~the~3D space, which is more in~line with the requirements of autonomous driving. In~addition, the~current mainstream application scenario of 3D object detection is autonomous driving, so that the research on object detection in~the~field of autonomous driving in~the~future will focus on 3D object detection. A~common method for vision-based 3D object detection is to estimate the key points of the 3D bounding box within a~2D bounding box. Such works include SSD-6D~\cite{094}, which is an~SSD-based six-dimensional pose estimation network, and~Mono3D~\cite{023}, which uses semantic and instance segmentation information to score candidate~boxes. Ref. \cite{023} also chose two-stage 3D object recognition, and its main contribution is to propose a disentangling transformation to separate multiple parameters that need to be trained to improve the efficiency of training. Ref. \cite{098} used a 3D anchor constraint box to directly obtain a 3D constraint box by regression and the gap between the anchor constraint box, in which a network using different convolution weights at different positions of the feature map was used to improve the performance. There are also methods that do not require 2D object detection first. These methods transform image features into new representations. Ref. \cite{099} proposed orthographic feature transform to convert the general image feature map into a 3D voxel map, then performed normalization in the vertical dimension, converted it to a bird's-eye feature map and regressed the 3D parameters on the bird's-eye view.
	
	3D object detection based on multi-sensor fusion also adds radar input branches and information fusion module on the basis of vision-based object detection network. Ref. \cite{096} used a~scheme similar to feature-level fusion, first rendering the radar points into a~rectangular area through 2D detection and~then performing 3D detection. In~addition, since lidar is rich in~features and can reconstruct object contours, it is easier to estimate 3D bounding boxes, there are more studies on multi-sensor fusion 3D object detection with lidar.}

\subsubsection{Lidar in~Autonomous~Driving}

As the cost of lidar decreases, autonomous driving vehicles equipped with lidar have become a~trend. However,~lidar is not a~substitute for mmWave radar. As~noted in~Section \ref{sec3}, mmWave radar has its own unique advantages. Lidar has higher detection accuracy, and~they complement each other's advantages. 
The fusion of lidar and vision is becoming promising in~autonomous driving. The~main research includes object detection, object classification, and road~segmentation.

\begin{itemize}
	\item Object Detection
\end{itemize}

The proposed fusion method in~\cite{081} first applied a lidar point cloud to generate suggestions for potential car positions in~the~image. The~position of the bounding box was then refined by mining multiple layers of information in~the~proposal network. Finally, the~object detection was performed by a~detection network that shared part of the layers with the proposed network. Ref.~\cite{084} proposed a~fusion method similar to the data level fusion of mmWave radar and vision. It utilized lidar instead of mmWave radar to generate initial recommendations for the position of objects. Multi-scale features were then exploited to realize accurate detection of objects of different~sizes.

\begin{itemize}
	\item Object Classification
\end{itemize}

Ref.~\cite{082} presented a~lidar and vision fusion method for object classification that combines Deep-CNN (DCNN) and upsampling theory. It can inherit the advantages of the two methods and avoid the shortcomings of each method. This method up-sampled the lidar information and converted it into depth information, which would be fused with the vision image data and sent into the DCNN for training. The~experimental results showed that this method exhibits superior effectiveness and accuracy of object~classification.

\begin{itemize}
	\item Road Detection
\end{itemize}

{Ref.}
~\cite{083} proposed a~method to combine lidar point clouds and vision images for road detection. First, the~method of point alignment was used to project the lidar point cloud onto the vision image. The~vision image was then upsampled to obtain a~set of dense images containing spatially encoded information. Finally, they trained multiple fully convolutional neural networks (FCN) for road detection. The~proposed FCN was evaluated on KITTI and got a~MaxF score of 96.03\%, which is one of the top-performing methods so~far.

\subsubsection{Multimodal Information~Fusion}

The objects we see, the~sounds we hear, and~the smells we smell in~the~real world are all different modal information. Combining the multimodal information of the same thing can help AI better understand the thing. At~present, the~more popular research in~this field involves the association of pictures, voices, videos, and~texts, such as search~engines.

Therefore, whether it is mmWave radar or lidar, its sensing information is the same environmental information in~different modes. Radar sensing information and vision information are also information of different modalities. Considering radar vision fusion as multimodal information fusion, there may be a~better~solution.

In addition, in~the~field of autonomous driving, the~mmWave radar data provided by the dataset are post-processed data. However, from~the perspective of information conservation, the~amount of information contained in~the~post-processed radar data must be lost relative to the original data. If~the original radar detection data and vision images are regarded as two different modalities of sensing information to be fused, more abundant sensing information can be~obtained.

The challenge of multimodal information fusion is how to perfectly combine the information of different modalities and the noise they carry and~how to mine the relation information to assist the understanding of the same~thing.

\section{Conclusions}\label{sec7}

Object detection is one of the most important tasks for autonomous driving. In~this article, we provide an overview of mmWave radar and vision fusion for vehicle detection. First, we introduced the tasks, evaluation criteria, and datasets of autonomous driving. Second, we divided the mmWave radar and vision fusion process into three parts and the fusion algorithms were classified into three levels: data level, decision level, and feature level. Finally, 3D object detection, lidar vision fusion and multimodal information fusion, as promising technology in~the~future, is reviewed.


\begin{thebibliography}{100}
	 
\bibitem{001} Towler, J.; Bries, M.  Ros-military: Progress and promise. In Proceedings of the {Ground Vehicle Systems Engineering \& Technology Symposium {(GVSETS)}},  Novi, MI, USA, 7--9 August {2018}. 

\bibitem{002} Waymo. Journey.  Available online: 
\url{https://www.waymo.com/journey/} (accessed on 23 July  2021).

\bibitem{004} SAE On-Road Automated Vehicle Standards Committee. Taxonomy and definitions for terms related to on-road motor vehicle automated driving systems. \emph{SAE Stand. J.} \textbf{2014}, \emph{3016},  1--16.

\bibitem{003} Ren, K.; Wang, Q.; Wang, C. The security of autonomous driving: Threats, defenses, and future directions. \emph{Proc.  IEEE}   \textbf{2019},   \emph{3},   357--372.



\bibitem{005} Dalal, N.; Triggs, B. Histograms of oriented gradients for human detection.  In Proceedings of the   {IEEE Computer Society Conference on Computer Vision and Pattern Recognition ({CVPR}'05)}, San Diego, CA, USA,  20--25 June 2015; Volume 1, pp. 886--893.

\bibitem{006} Felzenszwalb, P.; McAllester, D.; Ramanan, D. A discriminatively trained, multiscale, deformable part model. In Proceedings of the  {IEEE Conference on Computer Vision and Pattern {Recognition}}, Anchorage, AK, USA, 23--28 June 2008;  pp. 1--8.

\bibitem{007} Redmon, J.; Divvala, S.; Girshick, R.; Farhadi, A.  You only look once: Unified, real-time object detection. In Proceedings of the  {IEEE Conference on Computer Vision and Pattern {Recognition}},  Anchorage, AK, USA, 23--28 June 2008; pp. 1--8.


\bibitem{008} Liu, W.; Anguelov, D.; Erhan, D.; Szegedy, C.; Reed, S.; Fu, C.Y.; Berg, A.C. SSD: Single shot multibox detector. In Proceedings of the {14th European Conference on Computer Vision {(ECCV 2016)}}, Amsterdam, The Netherlands,  11--14  October 2016; pp.~21--37.

\bibitem{009} Lin, T.Y.; Goyal, P.; Girshick, R.; He, K.; Doll´ar, P. Focal loss for dense object detection.  \emph{IEEE Trans. Pattern Anal. Mach. Intell.} \textbf{2020},   \emph{42},  318--327. 

\bibitem{010} Girshick, R.; Donahue, J.; Darrell, T.; Malik, J. Rich feature hierarchies for accurate object detection and semantic segmentation.  In Proceedings of the  {IEEE Conference on Computer Vision and Pattern {Recognition}}, Columbus, OH, USA, 23--28  June 2014; pp. 580–587.

\bibitem{011} He, K.; Zhang, X.; Ren, S.; Sun, J. Spatial pyramid pooling in~deep convolutional networks for visual recognition. \emph{IEEE Trans. Pattern Anal. Mach. Intell.} \textbf{2015}, \emph{37},   346–361.  

\bibitem{012} Ren, S.; He, K.; Girshick, R.; Sun, J. Faster r-cnn: Towards real-time object detection with region proposal networks. \emph{IEEE Trans. Pattern Anal. Mach. Intell.}  \textbf{2017}, \emph{39},   91–99.

\bibitem{013} Michaelis, C.; Mitzkus, B.; Geirhos, R. Benchmarking robustness in~object detection: Autonomous driving when winter is coming. \emph{arXiv}   \textbf{2019}, {arXiv:1907.07484}. 

\bibitem{014} Geirhos, R.; Temme, C.R.M.; Rauber, J. Generalisation in~humans and deep neural networks. \emph{arXiv} \textbf{2018}, {arXiv:1808.08750}.

\bibitem{015} Zhang, R.; Cao, S. Real-time human motion behavior detection via CNN using mmWave radar. \emph{IEEE Sensors Lett.} \textbf{2019}, \emph{3},  1--4. 

\bibitem{016} Yoneda, K.; Hashimoto, N.; Yanase, R. Vehicle Localization using 76GHz Omnidirectional Millimeter-Wave Radar for Winter Automated Driving. In Proceedings of the {IEEE Intelligent Vehicles {Symposium (IV)}}, Changshu, China,  26--30 June 2018; \mbox{pp. 971--977}.

\bibitem{017} Nagasaku, T.; Kogo, K.; Shinoda, H. 77 GHz Low-Cost Single-Chip Radar Sensor for Automotive Ground Speed Detection.  In Proceedings of the  {IEEE Compound Semiconductor Integrated Circuits {Symposium}}, Monterey, CA, USA, 12--15 October 2008; pp. 1--4.

\bibitem{018} {Hines, M.E.; Zelubowski, S.A. Conditions affecting the accuracy of speed measurements by low power MM-wave CW Doppler radar.  In Proceedings of the  {Vehicular Technology Society 42nd VTS Conference-Frontiers of {Technology}}, Denver, CO, USA, 10--13 May  1992; Volume 2, pp. 1046--1050.}

\bibitem{019} Zou, Z.; Shi, Z.; Guo, Y. Object detection in~20 years: A~survey. \emph{arXiv}   \textbf{2019}, {arXiv:1905.05055}.

\bibitem{020} Ziebinski, A.; Cupek, R.; Erdogan, H. A survey of ADAS technologies for the future perspective of sensor fusion. In Proceedings of the  {International Conference on Computational Collective {Intelligence}},  Halkidiki, Greece, 28--30 September  2016; Volume 9876, pp. 135--146.

\bibitem{021} Wang, Z.; Wu, Y.; Niu, Q. Multi-sensor fusion in~automated driving: A~survey. \emph{IEEE Access} \textbf{2019},  \emph{8},  2847--2868. 

\bibitem{025} Geiger, A.; Lenz, P.; Urtasun, R. Are we ready for autonomous driving? The~KITTI vision benchmark suite. In Proceedings of the {IEEE Conference on Computer Vision and Pattern {Recognition}}, Providence, RI, USA, 16--21 June  2012; pp. 3354--3361.

\bibitem{022} Everingham, M.; Gool, L.V.; Williams, C.K.I.; Winn, J.; Zisserman, A. The PASCAL Visual Object Classes Challenge 2011 (VOC2011) Results. Available online: \url{http://host.robots.ox.ac.uk/pascal/VOC/voc2011/index.html} (accessed on 8 March 2022). 

\bibitem{093} The~KITTI Vision Benchmark Suite. 3D Object Detection Evaluation 2017. 
Available online: \url{http://www.cvlibs.net/datasets/kitti/eval\_object.php?obj\_benchmark=3d} (accessed on 16 February 2022). 

\bibitem{023} Simonelli, A.; Bulò, S.R.; Porzi, L. Disentangling Monocular 3D Object Detection. In Proceedings of the  {IEEE/CVF International Conference on Computer Vision {(ICCV)}},    Seoul, Korea,  27--28 October  2019; pp. 1991--1999.

\bibitem{024} Huang, X.; Cheng, X.; Geng, Q. The apolloscape dataset for autonomous driving. In Proceedings of the  {IEEE/CVF Conference on Computer Vision and Pattern Recognition Workshops ({CVPRW})}, Salt Lake City, UT, USA, 22--18 June  2018; pp. 1067--10676.



\bibitem{026} Cordts, M.; Omran, M.; Ramos, S. The cityscapes dataset for semantic urban scene understanding. In Proceedings of the {IEEE Conference on Computer Vision and Pattern Recognition ({CVPR})}, Las Vegas, NV, USA,   27--30 June 2016; pp. 3213--3223.

\bibitem{027} Sun, P.; Kretzschmar, H.; Dotiwalla, X. Scalability in~perception for autonomous driving: Waymo open dataset. In Proceedings of the {IEEE/CVF Conference on Computer Vision and Pattern Recognition ({CVPR})}, Seattle, WA, USA, 14--19 June 2020; pp. 2443--2451.

\bibitem{028} Caesar, H.; Bankiti, V.; Lang, A.H. nuScenes: A~multimodal dataset for autonomous driving.  In Proceedings of the {IEEE/CVF Conference on Computer Vision and Pattern Recognition ({CVPR})},  Seattle, WA, USA, 14--19 June  2020; p. 11618.

\bibitem{029} Tesla. Future of Driving.  Available online: \url{https://www.tesla.com/autopilot} (accessed on 13 July 2021). 

\bibitem{030} {Apollo. Robotaxi Autonomous Driving Solution. 
	Available online: \url{https://apollo.auto/robotaxi/index.html} (accessed on 23 July 2021).}

\bibitem{031} NIO. NIO Autonomous Driving. 
Available online: \url{https://www.nio.cn/nad} (accessed on 23 July 2021).

\bibitem{032} XPENG. XPILOT Driving. 
Available online: \url{https://www.xiaopeng.com/p7.html?fromto=gqad004} (accessed on 23 July 2021).

\bibitem{033} Audi. Audi AI Traffic Jam Pilot.  Available online: \url{https://www.audi-technology-portal.de/en/electrics-electronics/driver-assistant-systems/audi-a8-audi-ai-traffic-jam-pilot} (accessed on 23 July 2021).

\bibitem{034} Daimler. Drive Pilot.  Available online: \url{https://www.daimler.com/innovation/case/autonomous/drive-pilot-2.html} (\mbox{accessed} on 23~July 2021).

\bibitem{038} Cho, M. A Study on the Obstacle Recognition for Autonomous Driving RC Car Using LiDAR and Thermal Infrared Camera. In Proceedings of the  {Eleventh International Conference on Ubiquitous and Future Networks ({ICUFN})}, Zagreb, Croatia, 2-5 July 2019;  pp. 544--546.


\bibitem{035} Zhou, T.; Yang, M.; Jiang, K. MMW Radar-Based Technologies in~Autonomous Driving: A~Review. \emph{Sensors} \textbf{2020}, \emph{20},  7283. 

\bibitem{036} Dickmann, J.; Klappstein, J.; Hahn, M.; Appenrodt, N.; Bloecher, H.L.; Werber, K.; Sailer, A. Automotive radar the key technology for autonomous driving: From detection and ranging to environmental understanding. In Proceedings of the {IEEE Radar Conference (RadarConf)},  Philadelphia, PA, USA,  2--6 May  2016; pp. 1--6.

\bibitem{037} Dickmann, J.; Appenrodt, N.; Bloecher, H.L.; Brenk, C.; Hackbarth, T.; Hahn, M.; Klappstein, J.; Muntzinger, M.; Sailer, A. Radar contribution to highly automated driving. In Proceedings of the {2014 11th European Radar {Conference}}, Rome, Italy, 8–10 October 2014; pp. 412--415. 


\bibitem{039} Alland, S.; Stark, W.; Ali, M.; Hegde, M. Interference in~Automotive Radar Systems: Characteristics, Mitigation Techniques, and Current and Future Research. \emph{IEEE Signal Process. Mag.} \textbf{2019}, \emph{36},   45--59.

\bibitem{040} Continental. Radars Autonomous Mobility. Available online: \url{https://www.continental-automotive.com/en-gl/Passenger-Cars/Autonomous-Mobility/Enablers/Radars}  (accessed on 23 July 2021).

\bibitem{041} Milch, S.; Behrens, M. Pedestrian detection with radar and computer vision. In Proceedings of the {PAL 2001—Progress in~Automobile {Lighting}}, Darmstadt, Germany, 25--26 September 2001; Volume 9, pp. 657--664.



\bibitem{051} Huang, W.; Zhang, Z.; Li, W.; Tian, J. Moving object tracking based on millimeter-wave radar and vision sensor. \emph{J. Appl. Sci. Eng.} \textbf{2018},  \emph{21},   609--614. 

\bibitem{062} Streubel, R.; Yang, B. Fusion of stereo camera and MIMO-FMCW radar for pedestrian tracking in~indoor environments. In Proceedings of the  {2016 IEEE International Conference on Information {Fusion}}, Heidelberg, Germany,  5-8 July 2016; pp. 565–572.

\bibitem{049} Guo, X.; Du, J.; Gao, J.; Wang, W. Pedestrian Detection Based on Fusion of Millimeter Wave Radar and Vision.  In Proceedings of the {2018 International Conference on Artificial Intelligence and Pattern {Recognition}}, Beijing, China, 18-20  August 2018;  pp. 38–42.
\bibitem{077} Mo, C.; Yi, L.; Zheng, L.; Ren, Y.; Wang, K.; Li, Y.; Xiong, Z. Obstacles detection based on millimetre-wave radar and image fusion technique.  In Proceedings of the  IET International Conference on Intelligent and Connected Vehicles, Chongqing, China, 22--23 September 2016.
\bibitem{079} Bi, X.; Tan, B.; Xu, Z.; Huang, L. \emph{A New Method of Target Detection Based on Autonomous Radar and Camera Data Fusion}; SAE Technical Paper; SAE: Warrendale, PA, USA,  {2017}
.
\bibitem{061} Steux, B.; Laurgeau, C.; Salesse, L.; Wautier, D. Fade: A~vehicle detection and tracking system featuring monocular color vision and radar data fusion. In Proceedings of the {2002 IEEE Intelligent Vehicles {Symposium}}, Versailles, France,  17-21 June 2002; Volume 2, pp. 632–639.

\bibitem{085} Liu, F.; Sparbert, J.; Stiller, C. IMMPDA vehicle tracking system using asynchronous sensor fusion of radar and vision. In Proceedings of the  {2008 IEEE Intelligent Vehicles {Symposium}}, Eindhoven,  The Netherlands, 4-6  June 2008; pp. 168--173.
\bibitem{076} Alencar, F.; Rosero, L.; Filho, C.; Osório, F.; Wolf, D. Fast Metric Tracking by Detection System: Radar Blob and Camera Fusion. In Proceedings of the  {2015 Latin American Robotics Symposium and Brazilian Symposium on Robotics ({LARS-SBR})},  Uberlândia, Brazil, 28 October--1 November 2015; pp. 120--125.
\bibitem{087} Richter, E.; Schubert, R.; Wanielik, G. Radar and vision based data fusion---Advanced filtering techniques for a~multi object vehicle tracking system. In Proceedings of the {2008 IEEE Intelligent Vehicles {Symposium}}, Eindhoven,  The Netherlands, 4-6  June 2008;  pp. 120--125.



\bibitem{050} Kato, T.; Ninomiya, Y.; Masaki, I. An obstacle detection method by fusion of radar and motion stereo. \emph{IEEE Trans. Intell. Transp. Syst.} 2002, 3,  182--188. 
\bibitem{048} Wang, T.; Zheng, N.; Xin, J.; Ma, Z. Integrating millimeter wave radar with a~monocular vision sensor for on-road obstacle detection applications. \emph{Sensors} \textbf{2011}, \emph{11},  8992--9008.
\bibitem{071} Garcia, F.; Cerri, P.; Broggi, A.; Escalera, A.; Armingol, J.M. Data fusion for overtaking vehicle detection based on radar and optical flow. In Proceedings of the {2012 IEEE Intelligent Vehicles {Symposium}}, Madrid, Spain,  3-7  June 2012; pp. 494--499.
\bibitem{075} Wang, X.; Xu, L.; Sun, H.; Xin, J.; Zheng, N. Bionic vision inspired on-road obstacle detection and tracking using radar and visual information. In Proceedings of the {17th International IEEE Conference on Intelligent Transportation Systems ({ITSC})}, Qingdao, China,  8--11  October  2014; pp. 39--44.



\bibitem{045} Haselhoff, A.; Kummert, A.; Schneider, G. Radar-vision fusion for vehicle detection by means of improved haar-like feature and adaboost approach. In Proceedings of the   {IEEE European Signal Processing {Conference}}, Poznan, Poland, 3-7 September 2007; pp. 2070–2074.
\bibitem{074} Wang, T.; Xin, J.; Zheng, N. A Method Integrating Human Visual Attention and Consciousness of Radar and Vision Fusion for Autonomous Vehicle Navigation. In Proceedings of the  {2011 IEEE Fourth International Conference on Space Mission Challenges for Information {Technology}},   Palo Alto, CA, USA, 2-4  August 2011; pp. 192--197.






\bibitem{042} Bombini, L.; Cerri, P.; Medici, P.; Alessandretti, G. Radar-vision fusion for vehicle detection.  In Proceedings of the {International Workshop on Intelligent {Transportation}},    Hamburg, Germany, 14–15 March {2006}; Volume {65}, p. 70.







\bibitem{078} Wang, X.; Xu, L.; Sun, H.; Xin, J.; Zheng, N. On-Road Vehicle Detection and Tracking Using MMW Radar and Monovision Fusion. \emph{IEEE Trans. Intell. Transp. Syst.} \textbf{2016}, \emph{17}, 2075--2084. 




\bibitem{070} Tan, Y.; Han, F.; Ibrahim, F. A Radar Guided Vision System for Vehicle Validation and Vehicle Motion Characterization. In Proceedings of the  {2007 IEEE Intelligent Transportation Systems {Conference}},  Bellevue, WA, USA,  30 September--3 October 2007; pp. 1059--1066.

\bibitem{072} Yang, J.; Lu, Z.G.; Guo, Y.K. Target Recognition and Tracking based on Data Fusion of Radar and Infrared Image Sensors. In Proceedings of the {International Conference on Information {Fusion}}, Sunnyvale, CA, USA, 6--8 July 1999; pp. 6--8.
\bibitem{046} Ji, Z.P.; Prokhorov, D.  Radar-vision fusion for object classificatio.  In Proceedings of the {IEEE International Conference on Information {Fusion}},   Cologne, Germany, 30 June--3 July 2008; pp. 1–7.
\bibitem{043} Alessandretti, G.; Broggi, A.; Cerri, P. Vehicle and guard rail detection using radar and vision data fusion.   \emph{IEEE Trans. Intell. Transp. Syst.}  2007, 8,  95--105.



\bibitem{044}  Kadow, U.; Schneider, G.; Vukotich, A. Radar-vision based vehicle recognition with evolutionary optimized and boosted features. In Proceedings of the {IEEE Intelligent Vehicles {Symposium}},  Istanbul, Turkey, 13--15 June 2007; pp. 749–754. 










\bibitem{052} Langer, D.; Jochem, T. Fusing radar and vision for detecting, classifying and avoiding roadway obstacles. In Proceedings of the  {IEEE Conference on Intelligent {Vehicles}}, Tokyo, Japan, 19--20 September 1996; pp. 333–338.
\bibitem{058} Chavez, R.O.; Burlet, J.; Vu, T.D.; Aycard, O. Frontal object perception using radar and mono-vision. In Proceedings of the {2012 IEEE Intelligent Vehicles {Symposium}},  Madrid, Spain,  3-7 June 2012; pp. 159--164.

\bibitem{063} Long, N.; Wang, K.; Cheng, R.; Yang, K.; Bai, J. Fusion of millimeter wave radar and RGB-depth sensors for assisted navigation of the visually impaired. In Proceedings of the {Millimeter Wave and Terahertz Sensors and {Technology XI}}, Berlin, Germany, 5 October 2018; Volume 10800, pp. 21--28.

\bibitem{064} Long, N.; Wang, K.; Cheng, R.; Hu, W.; Yang, K. Unifying obstacle detection, recognition, and fusion based on millimeter wave radar and RGB-depth sensors for the visually impaired. \emph{Rev. Sci. Instrum.}   \textbf{2019}, \emph{{90}
}, 044102.

\bibitem{088} Wang, J.G.; Chen, S.J.; Zhou, L.B.; Wan, K.W.; Yau, W.Y. Vehicle Detection and Width Estimation in~Rain by Fusing Radar and Vision. In Proceedings of the {2018 International Conference on Control, Automation, Robotics and Vision ({ICARCV})}, Singapore, 18-21 Novemebr 2018;  pp. 1063--1068.

\bibitem{089} Jha, H.; Lodhi, V.; Chakravarty, D. Object Detection and Identification Using Vision and Radar Data Fusion System for Ground-Based Navigation.   In Proceedings of the {2019 International Conference on Signal Processing and Integrated Networks ({SPIN})}, Noida, India, 7-8  Mach 2019; pp. 590--593.

\bibitem{090} Kim, J.; Emeršič, Ž.; Han, D.S. Vehicle Path Prediction based on Radar and Vision Sensor Fusion for Safe Lane Changing. In Proceedings of the {2019 International Conference on Artificial Intelligence in~Information and Communication ({ICAIIC})}, Okinawa, Japan, 11--13 February 2019; pp. 267--271.







\bibitem{053} Coué, C.; Fraichard, T.; Bessiere, P.; Mazer, E. Multi-sensor data fusion using Bayesian programming: An automotive application. In Proceedings of the {2002 IEEE Intelligent Vehicles {Symposium}}, Madrid, Spain,  3-7 June 2002; Volume 2, pp. 442–447.

\bibitem{054} Kawasaki, N.; Kiencke, U. Standard platform for sensor fusion on advanced driver assistance system using bayesian network. In Proceedings of the {2004 IEEE Intelligent Vehicles {Symposium}},  Parma, Italy, 14-17 June 2004; pp. 250–255.

\bibitem{055} Ćesić, J.; Marković, I.; Cvišić, I.; Petrović, I. Radar and stereo vision fusion for multitarget tracking on the special Euclidean group. \emph{Robot. Auton. Syst.}  2016, 83,  338--348.
\bibitem{059} Zhong, Z.; Liu, S.; Mathew, M.; Dubey, A. Camera radar fusion for increased reliability in~ADAS applications. \emph{Electron. Imaging Auton. Veh. Mach.}  2018, 17, 258-1--258-4.

\bibitem{060} Kim, D.Y.; Jeon, M. Data fusion of radar and image measurements for multi-object tracking via Kalman filtering. \emph{Inf. Sci.} \textbf{2014}, \emph{278},  641--652.

\bibitem{056} Obrvan, M.; Ćesić, J.; Petrović, I. Appearance based vehicle detection by radar-stereo vision integration. In Proceedings of the {Robot 2015: Second Iberian Robotics {Conference}},  Lisbon, Portugal, 19-21 November 2015;  Volume 417, pp. 437–449.

\bibitem{057} Wu, S.; Decker, S.; Chang, P.; Camus, T.; Eledath, J. Collision sensing by stereo vision and radar sensor fusion.  \emph{IEEE Trans. Intell. Transp. Syst.} 2009,  10, 606-614.



















\bibitem{065} Chadwick, S.; Maddern, W.; Newman, P. Distant Vehicle Detection Using Radar and Vision. In Proceedings of the {2019 International Conference on Robotics and Automation ({ICRA})},   Montreal, QC, Canada, 20--24 May 2019; pp. 8311--8317.

\bibitem{066} John, V.; Mita, S. RVNet: Deep Sensor Fusion of Monocular Camera and Radar for Image-Based Obstacle Detection in~Challenging Environment. In Proceedings of the  {Pacific-Rim Symposium on Image and Video {Technology}, Sydney, Australia, 18--22 November 2019;  Volume 11854, pp. 351--364.}

\bibitem{067} Nobis, F.; Geisslinger, M.; Weber, M.; Betz, J.; Lienkamp, M. A Deep Learning-based Radar and Camera Sensor Fusion Architecture for Object Detection. In Proceedings of the {2019 Sensor Data Fusion: Trends, Solutions, Applications ({SDF 2019})}, Bonn, Germany, 15-17 October {2019}; pp. 1--7.



\bibitem{069} Chang, S.; Zhang, Y.; Zhang, F.; Zhao, X.; Huang, S.; Feng, Z.; Wei, Z. Spatial Attention Fusion for Obstacle Detection Using MmWave Radar and Vision Senso. \emph{Sensors} \textbf{2020}, \emph{20},  956.  
\bibitem{068} Lekic, V.; Babic, Z. Automotive radar and camera fusion using Generative Adversarial Networks.  \emph{Comput. Vis. Image Underst.} \textbf{2019}, \emph{184},  1--8.   


\bibitem{086} {Bertozzi, M.; Bombini, L.; Cerri, P.}; Medici, P.; Antonello, P.C.; Miglietta, M. Obstacle detection and classification fusing radar and vision. In Proceedings of the  {2008 IEEE Intelligent Vehicles {Symposium}},  Eindhoven, The Netherlands, 4--6 June 2008; pp. 608--613.

\bibitem{080} Hyun, E.; Jin, Y.S. Multi-level Fusion Scheme for Target Classification using Camera and Radar Sensors. In Proceedings of the {International Conference on Image Processing, Computer Vision, and Pattern Recognition ({IPCV})},  Las Vegas, NV, USA, 25--28 July 2017; pp. 111--114.
\bibitem{091} Simonyan, K.; Zisserman, A. Very deep convolutional networks for large-scale image recognition. \emph{arXiv} {\bf 2014},  {arXiv:1409.1556},
\bibitem{092} Tian, Z.; Shen, C.; Chen, H.; He, T. FCOS: Fully Convolutional One-Stage Object Detection. In Proceedings of the {2019 IEEE/CVF International Conference on Computer Vision ({ICCV})}, Seoul, Korea, 27 October--2 November 2019;  pp.  9627--9636.
\bibitem{094} Kehl, W.; Manhardt, F.; Tombari, F. Ssd-6d: Making rgb-based 3d detection and 6d pose estimation great again. In Proceedings of the {IEEE International Conference on Computer Vision ({ICCV})},  Venice, Italy, 22--29 October  2017; pp. 1521--1529.


\bibitem{098} Brazil G, Liu X. M3d-rpn: Monocular 3d region proposal network for object detection. In Proceedings of the IEEE/CVF International Conference on Computer Vision, Seoul, Korea, 27--28 October  2019; pp. 9287--9296.

\bibitem{099} Roddick T, Kendall A, Cipolla R. Orthographic feature transform for monocular 3d object detection. \emph{arXiv} \textbf{2018}, arXiv:1811.08188.


\bibitem{096} Nabati, R.; Qi, H. CenterFusion: Center-Based Radar and Camera Fusion for 3D Object Detection.  In Proceedings of the {IEEE/CVF Winter Conference on Applications of Computer Vision ({WACV})}, Waikoloa, HI, USA,  3-8 January 2021; pp. 1527--1536.
\bibitem{081} Du, X.; Ang, M.H.; Rus, D. Car detection for autonomous vehicle: LIDAR and vision fusion approach through deep learning framework.  In Proceedings of the {2017 IEEE/RSJ International Conference on Intelligent Robots and Systems ({IROS})}, Vancouver, BC, Canada, 24--28 September 2017;  pp. 749--754.
\bibitem{084} Zhao, X.; Sun, P.; Xu, Z.; Min, H.; Yu, H. Fusion of 3D LIDAR and Camera Data for Object Detection in~Autonomous Vehicle Applications. \emph{IEEE Sensors J.} \textbf{2020}, \emph{20},  4901--4913.  
\bibitem{082} Gao, H.; Cheng, B.; Wang, J.; Li, K.; Zhao, J.; Li, D. Object Classification Using CNN-Based Fusion of Vision and LIDAR in~Autonomous Vehicle Environment. \emph{IEEE Trans. Ind. Inform.}  \textbf{2018}, \emph{{14}, 4224 - 4231
}.

\bibitem{083} Caltagirone, L.; Bellone, M.; Svensson, L.; Wahde, M. LIDAR–camera fusion for road detection using fully convolutional neural networks. \emph{Robot. Auton. Syst.} \textbf{2019}, \emph{111},  125--131. 
	
	
\end{thebibliography}
\end{document}